\documentclass[stsy]{informs-stsy}

\OneAndAHalfSpacedXI
\usepackage{natbib}
 \bibpunct[, ]{(}{)}{,}{a}{}{,}%
 \def\bibfont{\small}%

\usepackage{MnSymbol}
\usepackage[inline]{enumitem}
\usepackage{algpseudocode}
\usepackage{verbatim}
\usepackage{bigstrut}
\usepackage{booktabs,multirow}
\usepackage{lscape}
\usepackage{graphicx}
\usepackage{float}
\usepackage{algorithm}
\usepackage{subfig}
\usepackage{mathtools}
\usepackage{pdfpages}
\TheoremsNumberedThrough     % Preferred (Theorem 1, Lemma 1, Theorem 2)
\ECRepeatTheorems

\EquationsNumberedThrough    % Default: (1), (2), ...
\MANUSCRIPTNO{MS-0001-1922.65}

\begin{document}
%%%%%%%%%%%%%%%%

% \setcounter{page}{1}
%\DOI{10.1287/xxxx.0000.0000}%

\RUNAUTHOR{Xiangyu and Peter}

\RUNTITLE{Near-optimality for infinite-horizon restless bandits with many arms}
\RUNTITLE{Infinite-horizon restless bandits with many arms}

% \TITLE{Bundling Information Goods of Decreasing Value}
\TITLE{Near-optimality for infinite-horizon restless bandits with many arms}

\ARTICLEAUTHORS{%
\AUTHOR{Xiangyu Zhang}
\AFF{Cornell University, \EMAIL{xz556@cornell.edu}} %, \URL{}}
\AUTHOR{Peter I. Frazier}
\AFF{Cornell University \EMAIL{pf98@cornell.edu}}
} % end of the block

\ABSTRACT{
Restless bandits are an important class of problems with applications in recommender systems, active learning, revenue management and other areas. 
We consider infinite-horizon discounted restless bandits with many arms where a fixed proportion of arms may be pulled in each period and where arms share a finite state space.
Although an average-case-optimal policy can be computed via stochastic dynamic programming, the computation required grows exponentially with the number of arms $N$. Thus, it is important to find scalable policies that can be computed efficiently for large $N$
and that are near optimal in this regime, in the sense that the optimality gap (i.e. the loss of expected performance against an optimal policy) per arm vanishes for large $N$.
However, the most popular approach, the Whittle index, requires a hard-to-verify indexability condition to be well-defined and another hard-to-verify condition to guarantee a $o(N)$ optimality gap. 
We present a method resolving these difficulties. By replacing a global Lagrange multiplier used by the Whittle index with a sequence of Lagrangian multipliers, one per time period up to a finite truncation point, we derive a class of policies, called fluid-balance policies, that have a $O(\sqrt{N})$ optimality gap. Unlike the Whittle index, fluid-balance policies do not require indexability to be well-defined and their $O(\sqrt{N})$ optimality gap bound holds universally without sufficient conditions. We also demonstrate empirically that fluid-balance policies provide state-of-the-art performance on specific problems.
}

\KEYWORDS{restless bandit, Markov decision processes, Whittle index} 
\maketitle
%%%%%%%%%%%%%%%%%%%%%%%%%%%%%%%%%%%%%%%%%%%%%%%%%%%%%%%%%%%%%%%%%%%%%%

\section{Introduction}
We study a stochastic control problem called the infinite-horizon restless bandit. In this problem, a decision maker is responsible for managing $N$ Markov decision processes (called ``arms'') whose states are fully observed and belong to a common finite state space.
For each arm in each time period, the decision maker can either activate the arm (also called ``pullng'' the arm) or idle it. 
This arm then generates a random reward. This reward's probability distribution depends in a known way on the action taken and the
arm's current state. 
A known transition kernel depending on the action and the arm's current state then determines the probability distribution over the arm's state in the next time period. When making decisions, the decision maker needs to respect a ``budget'' constraint in each period that constrains the number of arms that can be activated in each period.  The objective is to maximize the expected total discounted reward over an infinite time horizon.

The infinite-horizon restless bandit problem was first formulated by \cite{whittle1980multi} and has since attracted much theoretical and practical interest. 
Many real-world decision-making problems are naturally formulated as restless bandits, with applications arising in network communication \citep{liu2008restless}, unmanned aerial vehicles tracking \citep{le2006multi}, revenue management \citep{brown2020index} and active learning \citep{chen2013optimistic}.

In principle, the problem can be solved by value iteration or other standard methods for maximizing the infinite-horizon expected total discounted reward in a stochastic dynamic program \citep{powell2007approximate}. Unfortunately, the dimension of the collective description of arms' states grows linearly with $N$. Thus, the computation required grows exponentially with respect to $N$ due to the ``curse of dimensionality'' \citep{powell2007approximate}. 

Because optimality appears unachievable by computationally tractable algorithms, theoretical analysis of restless bandits has focused on {\it asymptotic optimality} for the asymptotic regime where the budget constraint grows proportionally with $N$.
% First proposed by \cite{whittle1980multi}, 
% In particular, following \cite{whittle1980multi}, people are interested in the asymptotic regime where budget constraint grows proportionally with $N$ in each period.
First defined by \cite{whittle1980multi}, an asymptotically optimal policy is one whose {\it optimality gap} (the difference between the given strategy's expected performance and that of an optimal policy, briefly, {\it opt gap}) divided by the number of arms vanishes as the number of arms grows.
% [Xiangyu to define {\it opt gap}, the difference in performance between the optimal policy and an approximate policy]

Asymptotic optimality has been hard to guarantee. The most popular approach to restless bandits, the Whittle index, was conjectured to be asymptotically optimal by \cite{whittle1980multi}. However, \cite{weber1990index} shows that this is false: there are problems where the Whittle index fails to be asymptotically optimal.  
That work also shows that the Whittle index is asymptotically optimal, but only if a certain hard-to-verify sufficient condition is met: that a differential equation characterizing the dynamics of the Whittle index in a certain fluid limit has a globally stable equilibrium.
Moreover, for the Whittle index to be well defined, the problem must satisfy a so-called ``indexability'' condition, which may not be met and is hard to verify in practice. 
% Since the difficulty and importance of the problem, many approaches have been proposed but with limited theoretical performance guarantees. Among them, Whittle index is the most popular one. 
% It was conjectured by Whittle \cite{whittle1980multi} that the Whittle index would achieve what we refer to as {\it asymptotic optimality}: that the opt gap (the difference between a given strategy's expected performance and that of an optimal policy) divided by the number of arms vanishes vanishes as the number of arms grows.
% However, \cite{weber1990index} show that the Whittle index can fail to be asymptotically optimal.  
% Moreover, defining the Whittle index for a problem requires the problem to satisfy s so-called ``indexability'' condition, which may not be met and is hard to verify in practice. 
Another popular approach is simulation-based \citep{meshram2020simulation, nakhleh2021neurwin}. However, the simulation-based method from \cite{meshram2020simulation} and \cite{nakhleh2021neurwin} does not provide a theoretical guarantee on performance.
% , and approximation-based methods only provide constant-factor approximations guarantees and may fail to be asymptotically optimal.

Although the existing infinite-horizon restless bandit policies of which we are aware suffer from difficulty in guaranteeing asymptotic optimality and, in the case of the Whittle index, challenges in establishing indexability and coping with its absence, 
 some recent work shows that life is much easier for finite-horizon restless bandits. 
 
For example, for finite-horizon restless bandits in the same asymptotic regime where budgets grow proportionally with $N$, \cite{hu2017asymptotically} propose a policy with $o(N)$ opt gap, thus being asymptotically optimal. Later, \cite{brown2020index} proposes policies with stronger $O(\sqrt{N})$ opt gaps. Moreover, \cite{zhang2021restless} propose a class of policies with at most $O(\sqrt{N})$ opt gaps and surprisingly $O(1)$ opt gap if a non-degeneracy condition is met.
These rates for opt gaps hold universally, unlike the hard-to-verify conditions required for the Whittle index to be asymptotically optimal.
Moreover, neither policy requires an indexability condition. 
These papers overcome the challenges articulated above despite basing their analysis on the same Lagrangian relaxation technique proposed and used by 
\cite{whittle1980multi}.

We argue in this paper that a key difference in the approach enabled these finite-horizon analyses to achieve asymptotic optimality and to avoid challenges in establishing indexability: their Lagrangian relaxation uses a {\it sequence} of Lagrange multipliers, one for each time period, while the Whittle index uses a single global Lagrange multiplier.
Using a time-varying Lagrange multiplier is intuitive in the finite-horizon setting:
the finite horizon causes the problem to be non-stationary, naturally inspiring a time-inhomogeneous approach.

We show in this paper that this time-inhomogeneous approach can be generalized to the infinite-horizon setting to overcome the shortcomings of the Whittle index and other past approaches to the infinite-horizon restless bandit.
That a time-inhomogeneous approach would be relevant to the infinite-horizon setting may, at first glance, seem surprising: the infinite-horizon problem is stationary, implying the existence of stationary optimal policies, and suggesting that asymptotically optimal policies should also be stationary.
Part of our contribution is to explain why non-stationarity is an important tool for providing asymptotic optimality in stationary infinite-horizon problems. 

We provide a novel class of computationally scalable non-stationary infinite-horizon restless bandit policies called ``fluid-balance'' policies. We show that they are asymptotically optimal, achieving a $O(\sqrt{N})$ opt gap. This result does not require indexability or other sufficient conditions beyond those defining the problem we study, such as arms' states belonging to a finite state space and state transitions that are conditionally independent across arms.
Moreover, despite being time-inhomogeneous in an infinite horizon problem, we show that they can be computed in finite time.
They are computed by considering a finite linear program formed by truncating the infinite-horizon problem. Truncating at the $O(\log N)$-th period allows fluid-balance policies to achieve a $O(\sqrt{N})$ opt gap.

This requires going substantially beyond applying a previously proposed finite-horizon policy to the truncated problem. 
Policies with $O(\sqrt{N})$ opt gaps in the finite-horizon setting proposed in \cite{brown2020index} and \cite{zhang2021restless} 
% Policies with asymptotic optimality in finite-horizon settings proposed in \cite{hu2017asymptotically,zayas2019asymptotically,brown2020index,zhang2021restless} 
have opt gap bounds that depend exponentially on the time horizon. Simply applying one of these policies and its associated performance bound to a truncated problem (and leveraging discounting to bound the reward obtained after truncation) results in an opt gap bound that grows faster than $\sqrt{N}$.
% , even if the truncation point is chosen carefully in the associated analysis.
Our fluid-balance policies and their analysis are 
specifically adapted to the infinite-horizon setting to circumvent this challenge.
% specifically adapted to the infinite-horizon setting:
% we show that, under these policies, the deviation of the occupation measure between the stochastic and fluid limit problems is bounded over all time periods, which in turn implies a bound on a single period's contribution to the opt gap.

% An analysis of the occupation measure is also key to avoiding the exponential dependence on the time horizon that plagues 
% we show that, under these policies, the deviation of the occupation measure between the stochastic and fluid limit problems is bounded over all time periods, which in turn implies a bound on a single period's contribution to the opt gap.

We give intuition for why a non-stationary approach can resolve the past challenges in infinite-horizon restless bandits.
Index policies, such as the Whittle index and the policies that we propose, operate by defining a priority or ``index'' for each state and then pulling arms in order from the ones in the highest priority to the lowest priority until the budget constraint on pulls in the current time period is exhausted.
Essentially, non-stationary approaches are beneficial because performing well with an index policy requires using a different priority order over states in each time period.

Understanding the need to use a time-dependent priority order
relies on a linear programming analysis of the so-called fluid approximation. In this approximation, we take the limit as the number of arms grows large while scaling up the budget constraint. 
A policy can be understood as taking an occupation measure (a vector comprising the fraction of arms in each state at a particular time) as input and deciding the fraction of arms in each state to pull.
In the fluid limit, an optimal policy's decisions in a time period can be understood as pulling arms according to their marginal benefit from high to low until all resources are consumed. 
This marginal benefit in each period depends on the occupation measure in that period.
Critically, under an optimal policy in the fluid limit, the occupation measure changes over time. This causes  the optimal ranking over states to vary across periods.
% In the fluid limit, under continuity conditions on the policy, the occupation measure at each time $t$ becomes deterministic. 
% Moreover, one can compute a policy that is optimal using a linear program.
% Note to self: the proof of this is that the LP in the fluid limit is equivalent to one where we lagrangify the budget constraint, and the Lagrangian is separable across arms 
% The marginal benefit includes both the change in immediate reward due to the pull and the change in future value induced by altering the arm's stochastic transition in state.
Thus, matching the optimal policy in the fluid limit requires an index policy to use a different a priority order in different time periods.

% This explains why any static state-priority policy does not seem to work in all circumstance thus requires the indexability condition. However, by applying the technique of time-adaptive penalties, we have the flexibility to handle this distribution shift by prioritizing states dynamically. Thus, a non-stationary policy (in the sense of priority over states) could perform much better than all stationary policies (in the sense of priority over states).

The rest of the paper is structured as follows. First,
Section \ref{lit-review} discusses relevant past work on infinite-horizon restless bandits and the novelty of our work. Then, Section \ref{sys_model} formulates the restless bandit formally as a Markov decision process and Section \ref{approx} discusses a standard linear programming relaxation used to support analysis of restless bandits in the past literature. Our proposed fluid-balance policies and analysis are also based on this relaxation. Section \ref{sec-diff-balance} then introduces a technical condition, diffusion regularity that is sufficient for a $O(\sqrt{N})$ opt gap. Section \ref{sect-6} proposes our fluid-balance policies and shows that are diffusion regular and thus have a $O(\sqrt{N})$ opt gap. Section \ref{numerical} uses two numerical experiments to explore the performance of fluid-balance policies. Finally, Section \ref{conclusion} concludes the paper and discusses possible future work.

\section{Literature Review}\label{lit-review}
This section first reviews approaches specifically designed for the infinite-horizon setting. It then reviews recent progress in the finite-horizon setting motivating our approach.
% in later sections.
\subsection{Infinite-horizon restless bandits}
The infinite-horizon restless bandit problem was first formulated by \cite{whittle1980multi}. Since then, the problem has attracted substantial research interest, both from theoretical and practical perspectives. Here we review two main streams of this research: the Whittle index and simulation-based approaches.

\textit{Whittle index}
When the restless bandit problem was first formulated in \cite{whittle1980multi}, this paper also proposed an index policy, the so-called Whittle index, as a solution. 
The Whittle index is defined by considering a problem with a single arm in which one can pull the arm, paying a cost, or idle it. The Whittle index for a state is the cost that makes an optimal policy indifferent between pulling the arm and idling it. This implies a ranking over states that, intuitively, is the same as ranking by a state's ``marginal productivity'': the difference in discounted long-run reward between activating and idling an arm in this state in the original problem \citep{nino2007dynamic}.
% The Whittle index policy activates arms according to their indices, from high to low, until the budget constraint is met.
% In a linear programming relaxation of the restless bandit (formulated below) that relaxes the budget constraint to a penalty on pulling arms to create a separable collection of problems, one per arm, the Whittle index can be interpreted as the largest Lagrange multiplier on the budget constraint such that a fluid-optimal policy would be willing to pull the arm.
% Intuitively, pulling arms in the order of their Whittle indices until reaching the budget constraint should correspond to pulling those arms 
% This can be interpreted as 
% Whittle index would assign each arm an index representing the ``marginal productivity'': the difference of discounted long-run reward with respect to activating or idling the arm. 
Intuitively, it should be a good policy to simply pull the arms in the states with the highest marginal productivity. Then Whittle index policy does exactly this: it activates arms according to their indices, from high to low, until all resources are used.

Although intuitively promising, \cite{whittle1980multi} noticed that the willingness to pull an arm in a single-arm problem is not always monotone: it may be optimal to pull the arm when the cost-per-pull is low, idle it when the cost-per-pull is in an intermediate range, and pull it when the cost-per-pull is high. In such settings, the Whittle index is not well-defined and its link to marginal productivity is lost.
% \xzedit{where indexability guarantees that there exists a ranking over arms' states in their marginal productivity that does not depend on the number of arms in each state.}
% \xzedit{Whittle index seems theoretically promising since it essentially generalizes Gittin index \citep{gittins1979bandit}, which is shown to be optimal for the setting where arms do not change state when idled and only a single arm can be activated in each period. Thus,}
\cite{whittle1980multi} conjectured that indexability would imply asymptotic optimality: the difference between the Whittle index's expected performance and that of an optimal policy divided by the number of arms vanishes as the number of arms grows, allowing a constant fraction of the arms to be pulled per time period.
Later, however, \cite{weber1990index} provided a counterexample to Whittle's conjecture: the Whittle index can fail to be asymptotically optimal even when the indexability condition is satisfied. 

Responding to the challenge of establishing indexability,
\cite{gittins2011multi, nino2001restless} 
establish alternate sufficient conditions for indexability 
and \cite{glazebrook2006some}
characterize some indexable restless bandit families. 
\cite{liu2008restless, liu2010indexability} and \cite{le2008multi} show their studied system is indexable. 
\cite{guha2007approximation, guha2008sequential, guha2010approximation, ansell2003whittle} and \cite{jacko2007time} have extended these ideas to more general settings e.g. convex reward, convex resource budget, stochastic arriving and leaving arms, etc.
Nevertheless, establishing indexability remains challenging for most problems and typically entails additional theoretical work that must be done on a problem-by-problem basis. 

When a problem is not indexable, multiple values satisfy the conditions that usually define the Whittle index. Thus, attempting to deploy a Whittle index policy in practice without first verifying indexability requires the implementation to explicitly handle this non-uniqueness. The use of implementations assuming a unique Whittle index value in non-indexable problems creates a risk that Whittle index computation produces errors or fails to converge{.} % \sout{, or lead to poor performance.}
{Also, the intuition for why a Whittle index policy would perform well relies on indexability. When indexability is lacking, policies prioritizing arms based on a Whittle index computation (while handling non-uniqueness) may be less likely to perform well.}

If indexability can be verified, establishing asymptotic optimality requires verifying the additional sufficient conditions discussed above. Past literature suggests that this may be even more difficult than verifying indexability. Most work using Whittle indices does not prove its asymptotic optimality in the problem studied \citep{liu2008restless, le2008multi} or only proves it in a specific parameter regime \citep{liu2010indexability, verloop2016asymptotically}. Instead, past literature often relies on numerical simulation to justify the Whittle index's performance.

Thus, despite its popularity, the difficulty of verifying indexability and the additional conditions needed for asymptotic optimality remain a challenge when applying Whittle index policy in real-world problems.

\textit{Simulation-based approaches} Responding to the limitations of the Whittle index policy, simulation-based approaches have been developed. For example, \cite{meshram2020simulation} develop rollout-based heuristic policies 
and \cite{nakhleh2021neurwin} and \cite{wang2021learning} develop a deep reinforcement learning strategy, using neural networks to approximate the value function. Numerical performance on a collection of benchmark problem instances is their primary concern rather than theoretical guarantees. {A policy that performs well in the problem instances simulated may perform poorly in other closely-related problem instances, and so performing well in a simulation-based study may not guarantee good performance across a wider range of problem instances faced after a policy is deployed to the field.}

{
Moreover, if all benchmark policies included in a numerical study are asymptotically suboptimal, a new asymptotically optimal policy has the potential to significantly outperform all of them. Thus, identifying new asymptotically optimal policies is of significant interest.
}

% Question: what if $N$ is large 

% \textit{Heuristic approaches} Many heuristic policies are proposed for infinite-horizon problems. Some heuristic are based on specific problem structures: for example, \cite{wei2015power} propose a heuristic for wireless networking and \cite{almeida2020hyper} propose heuristics for shop-flow problems. Other heuristic are less problem specific, and are based on some form of relaxation of the original problem: for example, \cite{bertsimas2000restless} and \cite{guha2007approximation}. Whether problem specific or not, these heuristics are not guaranteed to achieve asymptotic optimality.

% \textit{Learning approaches} \red{Xiangyu: add more details}

\subsection{Finite-horizon restless bandits}
While the Whittle index faces challenges in verifying indexability and the additional conditions required for asymptotic optimality, recent progress on finite-horizon restless bandits provides algorithms without these drawbacks in this alternate setting.

In rapid succession,
\cite{hu2017asymptotically,zayas2019asymptotically,brown2020index}
proposed index policies and show that they have $o(N)$, $O(\sqrt{N} \log N)$ and $O(\sqrt{N})$ opt gaps respectively.
Then, \cite{zhang2021restless} proposed a class of index policies
generalizing \cite{brown2020index} and \cite{hu2017asymptotically}, showing that this larger class of policies have at most a $O(\sqrt{N})$ opt gap and, surprisingly, a $O(1)$ opt gap if a non-degeneracy condition is met. 

Unlike the Whittle index, these index policies do not require an indexability condition to be well-defined. Moreover, they come with performance guarantees that do not require verifying extra sufficient conditions: $O(\sqrt{N})$ for \citealt{zhang2021restless,brown2020index,hu2017asymptotically},
$O(\sqrt{N} \log N)$ for \citealt{zayas2019asymptotically}.

We build on ideas in these finite-horizon papers to develop policies and analysis for the infinite-horizon setting that avoids the drawbacks of past infinite-horizon work: we develop policies that are asymptotically optimal and do not require indexability or other sufficient conditions.

{
This requires substantial additional analysis. One might hope to simply truncate the infinite-horizon problem, apply a previously proposed finite-horizon policy with a $O(\sqrt{N})$ finite-horizon performance guarantee to this truncated problem, and choose the truncation point large enough that the reward obtained afterward is a small part of the overall reward. This, however, does not produce a guarantee of asymptotic optimality in the infinite-horizon setting. 
}

% \pfcomment{Xiangyu to refine this:
% Our approach is to truncate 
% One might consider simply 
% A simple attempt to use such a policy 
Indeed, 
previously proposed policies known to have $O(\sqrt{N})$ opt gap in the finite-horizon setting
have performance guarantees that depend exponentially on $T$: the opt gap for a problem with horizon $T$ is 
$O(\sqrt{N}) O(\exp(\alpha T))$ with $\alpha > 0$ for both \cite{brown2020index} (see its Proposition 5) and \cite{zhang2021restless} (see its proof of Lemma 5). Thus, applying either policy and its associated bound to an infinite-horizon discounted problem truncated at $T$ with discount factor $\gamma$ would have a bound of $O(\sqrt{N}) O(\exp(\alpha T))$ on the opt gap realized up to the truncation time and a bound of $O(N \gamma^T)$ on the opt gap realized after the truncation time. Choosing $T$ to minimize the sum of these bounds would not provide a $O(\sqrt{N})$ opt gap bound.

\section{System Model}\label{sys_model}
This section formulates the restless bandit problem as a Markov decision process (MDP).

\textbf{Model}
The decision maker faces $N$ arms. Each arm is as an MDP, which is associated with a state space and an action space. 
The arms share the same finite state space $S$ and the same binary action space $A = \{0, 1\}$. For arm $i$, we let $s_{t, i} \in S$ denote its state and $a_{t, i} \in A$ the action applied in period $t \in \{1, 2, 3, ... \}$.

As we move from time period $t$ to $t+1$, each arm $i$ transitions to its new state
$s_{t + 1, i}$ independently given its 
current state $s_{t, i}$ and the action applied $a_{t, i}$. 
We use a kernel to describe this stochastic transition. The kernel is assumed known to the decision maker and is denoted by $P = \{p(s, a,  s')\}_{s, s' \in S, a \in A}$ where $$p(s, a, s') := \mathbb{P}[s_{t + 1, i} = s' | s_{t, i} = s, a_{t, i} = a],$$
and $p(s, a, s')$ gives the probability of an arm transitioning to state $s'$ conditioned on its current state being $s$ and action $a$ being taken. We assume that each arm shares the same transition kernel and the transition kernel is time-homogeneous. Thus, as we write above, the transition kernel $P$ does not depend on the arm index $i$ or time index $t$. For simplicity, we assume each arm starts from a common state $s^*$ at $t =  1$.
Our analysis also applies if each arm's initial state is chosen independently from a common distribution.

The decision maker must pull $\lfloor \alpha N \rfloor$ arms in each period, which we refer as the budget constraint. The constant $\alpha$ is also known to the decision maker.

In each period, an arm generates a real-valued reward $r(s, a)$ that is a deterministic function of its state $s\in S$ and the action applied $a\in A$. 
The decision maker's objective is to maximize the total infinite-horizon expected discounted reward collected across all $N$ arms with discount factor $\gamma$ while respecting the budget constraint in each period. 

To formulate this $N$-arm decision-making problem as a MDP, we introduce some additional notation. These $N$ arms form a new MDP, which we refer as the joint MDP. The state space $\mathbb{S}$ of this joint MDP is the Cartesian product of $N$ single-arm state spaces $S$: $\mathbb{S} =  S \times S \times ... \times S$. Similarly, the action space $\mathbb{A}$ of the joint MDP is the Cartesian product of $N$ single-arm action spaces $A$: $\mathbb{A} =  A \times A \times ... \times A$. At period $t$, we denote the state of the joint MDP as $\textbf{s}_t = (s_{t, 1}, ..., s_{t, N})$ and the action of the joint MDP as $\textbf{a}_t = (a_{t, 1}, ..., a_{t, N})$, where $i$-th components of $\textbf{s}_t$ and $\textbf{a}_t$ refer respectively to the state of arm $i$ and the action applied to it.
% Only for element $\textbf{a} = (a_1, ..., a_N)$ in action space $\mathbb{A}$, we define its norm as $|\textbf{a}| := \sum_{i = 1}^N a_i$.

Since the state of each arm evolves independently given its previous state and the action applied, the probability of transitioning from one state to another in the joint MDP is the product of the each arm's transition probability. Mathematically speaking, 
\begin{align*}
    \mathbb{P}[\textbf{s}_{t + 1} | \textbf{s}_t, \textbf{a}_t] = \prod_{i = 1}^N \mathbb{P}[s_{t + 1, i} | s_{t, i}, a_{t, i}].
\end{align*}

To clearly describe the budget constraint, we introduce a norm in the state space $\mathbb{A}$. For an element $\textbf{a} = (a_1, a_2, ..., a_N) \in \mathbb{A}$, its norm $|\textbf{a}| := \sum_{i = 1}^N a_i$ is the sum of its components, noting that these components are non-negative. This norm $|\textbf{a}_t|$ gives the number of arms pulled in period $t$. Thus, we can write our budget constraint as $|\textbf{a}_t| = \lfloor \alpha N \rfloor$ for each period $t$.

The reward of the joint MDP is the sum of the rewards generated by each individual arm. To formalize this, we define the joint MDP's reward function $R: \mathbb{S} \times \mathbb{A} \rightarrow \mathbb{R}$ via $R(\textbf{s}_t, \textbf{a}_t) = \sum_{i = 1}^N r(s_{t, i}, a_{t, i})$.

A policy $\pi: \mathbb{S} \times \{1, 2, ...\} \rightarrow \mathbb{A}$ is a mapping from the product of the state space and set of possible times to the action space. Under a policy $\pi$, the action taken in period $t$ is $\textbf{a}_t = \pi(\textbf{s}_t, t)$. The decision maker's objective is to choose a policy $\pi$ that maximizes the joint MDP's total expected discounted reward while respecting budget constraints. Mathematically speaking, this is the following stochastic constrained optimization problem,
\begin{align}\label{OP}
\max_{\pi} \ &\mathbb{E}_{\pi} \sum_{t = 1}^{\infty} \gamma^t R(\textbf{s}_t, \textbf{a}_t) \nonumber \\
s.t. \ &|\textbf{a}_t| = \lfloor \alpha N \rfloor, \forall t;
\end{align}
where $\mathbb{E}_{\pi}$ takes the expectation under the distribution on states, actions, and rewards induced by the policy $\pi$.

To measure the performance of a policy $\pi$, we define its value function, $$V_N(\pi) := \mathbb{E}_{\pi} \sum_{t = 1}^{\infty} \gamma^t R(\textbf{s}_t, \textbf{a}_t),$$
as the expected total discounted reward collected by this policy. An optimal policy has performance $V_N^* := \sup_{\pi} V_N(\pi)$. We define the \textit{optimality gap (opt gap)} of a policy $\pi$ as 
$$V_N^* - V_N(\pi),$$
i.e., the difference in performance between this policy and an optimal policy.
The smaller the opt gap, the better the policy.

As a MDP with a large but finite state space, Problem \ref{OP} can be solved in principle via dynamic programming. However, the time complexity of this approach grows exponentially with the number of arms $N$ because of the so-called curse of dimensionality \citep{powell2007approximate}: the joint MDP has a state space whose cardinality is $|S|^N$. Thus, we would like to find policies that are both computationally tractable and have strong theoretical performance guarantees in the regime with many arms.

\section{Background: Preliminary Results and Notation}\label{approx}
This section describes a linear programming relaxation of the restless bandit problem. This is a standard technique from the restless bandit literature. It is not part of our contribution and we introduce it simply to provide a self-contained treatment of our research contribution. 

The relaxation provides an upper bound on an optimal policy's performance, which can in turn bound the opt gap of any feasible policy. Also, the relaxation can be solved efficiently and its solution will provide insights into the design of an asymptotically optimal policy in later sections.

\textbf{Linear Programming Relaxation} Following \citep{hu2017asymptotically, zayas2019asymptotically, brown2020index, zhang2021restless}, which apply linear programming relaxations to finite-horizon restless bandits, we describe an equivalent relaxation for the infinite-horizon problem. We emphasize that this is not part of our research contribution and is introduced so that we can define notation and so that our treatment can be self-contained.

Instead of solving the original problem (\ref{OP}) with cardinality constraints on the number of arms pulled in each period, we consider a modified version where these cardinality constraints are relaxed to constraints on the expected number of arms pulled:
\begin{align}\label{KP_1}
    \hat{V}_N := \max_{\pi} \ & \mathbb{E}_{\pi} \sum_{t = 1}^{\infty} \gamma^t R(\textbf{s}_t, \textbf{a}_t) \nonumber \\ 
    s.t. \ & \mathbb{E}_{\pi}|\textbf{a}_t| = \alpha N, \forall t.
\end{align}
From now on we assume $\alpha$ is a rational number and $N$ (an integer) is chosen such that $\alpha N$ is also an integer. Thus, we drop the floor operator in the constraint (\ref{KP_1}). When $\alpha N$ is not an integer, analysis in Appendix \ref{al:2-1} shows that the rounding error caused by the floor operator does not affect any asymptotic analysis in later sections.

To support computation, we consider a version of problem (\ref{KP_1}) that is truncated at some horizon $T \le \infty$, introducing approximation error discussed below. 
In some problems with special structure, truncation will be unnecessary for computation allowing us to choose $T=\infty$, while in others it will be necessary, requiring $T<\infty$.
% We define $T\le \infty$ to be this truncation point, where we allow $T$ to be infinity to support those cases in which \eqref{KP_1} can be solved exactly. 
% Unlike the relaxed finite-horizon problem in \cite{hu2017asymptotically}, the infinite-horizon problem (\ref{KP_1}) cannot usually be solved exactly using the techniques described below because they would require an infinite number of decision variables and constraints. Instead, one can solve a version of problem (\ref{KP_1}) that is truncated at some finite horizon, introducing approximation error discussed below. We define $T\le \infty$ to be this truncation point, where we allow $T$ to be infinity to support those cases in which \eqref{KP_1} can be solved exactly. 
We denote the truncated relaxed problem and truncated original problem as
\begin{align}\label{KP}
    \hat{V}_N(T) := \max_{\pi} \ & \mathbb{E}_{\pi} \sum_{t = 1}^{T} \gamma^t R(\textbf{s}_t, \textbf{a}_t) \nonumber \\ 
    s.t. \ & \mathbb{E}_{\pi}|\textbf{a}_t| = \alpha N, \forall t;
\end{align}
and 
\begin{align}\label{truncated-original}
    V^*_N(T) := \max_{\pi} \ & \mathbb{E}_{\pi} \sum_{t = 1}^{T} \gamma^t R(\textbf{s}_t, \textbf{a}_t) \nonumber \\
    s.t. \ & |\textbf{a}_t| = \alpha N, \forall t.
\end{align}

This truncated relaxed problem (regardless of the value of $T$) can be decomposed across arms by an analysis similar to Fenchel duality, allowing us to solve \eqref{KP} with $N$ arms via an equivalent single-arm problem.
Second, since the feasible policies for the truncated relaxed problem \eqref{KP} is a superset of the feasible policies for the truncated original problem \eqref{truncated-original}, the value of \eqref{KP} provides an upper bound on the value of \eqref{truncated-original}. We state these properties formally in Lemma \ref{fenchel}, whose proof is left to Appendix \ref{proof-fenchel}.
\begin{lemma}\label{fenchel}
    $V^*_N(T) \le \hat{V}_N(T) = N \ \hat{V}_1(T).$
\end{lemma}

The single-arm truncated relaxed problem $\hat{V}_1(T)$ can also be formulated as a linear program. Choosing the components of the occupation measure $x_t(s, a) := \mathbb{P}_{\pi}[s_t = s, a_t = a]$ as decision variables, the single-arm truncated relaxed problem can be formulated as
\begin{align}\label{LP}
    \max & \ \sum_{t = 1}^T \gamma^t \sum_{s, a} x_t(s, a) r(s, a)\\
    s.t. 
    & \ \sum_{a} x_{t + 1}(s', a) = \sum_{s, a} x_t(s, a) p(s, a, s'),  \forall t \le T - 1, s \in S; \nonumber\\
    & \ \sum_{s} x_t(s, 1) = \alpha, \forall t \le T;\nonumber \\
    & \ \sum_{a} x_1(s^*, a) = 1, \sum_{s, a} x_1(s, a) = 1, \sum_{s, a} x_t(s, a) \ge 0, \forall t \le T.\nonumber
\end{align}
The first constraint ensures flow balance in each time period. The second constraint ensures that the budget constraint on the expected number of arms pulled is satisfied in each period. The third, forth and fifth constraints ensure that the occupation measure forms a probability measure in each period. Denoting the solution to \eqref{LP} by $\{ x_t(s, a) \}_{t, s, s}$, we have $\hat{V}_1(T) = \sum_{t = 1}^T \gamma^t \sum_{s, a} x_t(s, a) r(s, a)$.

% Linear program \ref{LP} can be solved using business LP solver

\textbf{Additional Notation}
Starting in the next section, we analyze the optimal occupation measure and the number of arms in each state under a stochastic sample path. To support this analysis, we introduce some additional notations here.

Given an optimal occupation measure solving (\ref{LP}), we let $z_t(s) := \sum_{a} x_t(s, a)$ denote the probability that an arm being in state $s$ at period $t$. For notational simplicity, we use the vectors $z_t = (z_t(s))_{s}$ and $x_t = (x_t(s, a))_{s, a}$ to denote the distribution over an arm's state and the distribution over an arm's state-action pair given an optimal occupation measure.

We are also interested in the realized number of arms in each state under a stochastic sample path. We let $Z_t^N(s)$ denote the number of arms in state $s$ in period $t$ and let $X_t^N(s, a)$ denote the number of arms in state $s$ for which we took action $a$ taken in period $t$. We have that $Z_t^N(s) = \sum_{a} X_t^N(s, a)$ for any $t$ and $s$. Similar to the vector notation used to describe an optimal occupation measure, we use the vectors $Z_t^N = (Z_t^N(s))_{s}$ and $X_t^N = (X_t^N(s, a))_{s, a}$.

Starting from Section \ref{sec-diff-balance}, we will be interested in deviations between the realized number of arms under a stochastic sample path from an optimal occupation measure. To characterize this deviation, we define the following {\it diffusion statistics}:
\begin{align*}
    \tilde{Z}^N_t(s) = \frac{Z^N_t(s) - N z_t(s)}{\sqrt{N}}, \ \tilde{X}^N_t(s, a) = \frac{X^N_t(s, a) - N x_t(s, a)}{\sqrt{N}}.
\end{align*}
We also use the vectors $\tilde{Z}_t^N = (\tilde{Z}^N_t(s))_{s}$ and $\tilde{X}_t^N = (\tilde{X}^N_t(s, a))_{s, a}$ for notational simplicity.

With this new notation, we can rewrite a policy $\pi$ in term of its diffusion statistics.
Given a policy $\pi: \mathbb{S} \times \{1, 2, ... \} \rightarrow \mathbb{A}$, we can rewrite it as a sequence of mappings $\{ \tilde{\pi}_{t, N} \}_t$ s.t.
\begin{align*}
    \pi(Z^N_t, t) = X_t^N \iff	\tilde{\pi}_t^N(\tilde{Z}_t^N) = \tilde{X}_t^N.
\end{align*}
We refer to the sequence $\{ \tilde{\pi}_{t, N} \}_t$  as the {\it induced maps} from policy $\pi$. Section \ref{sec-diff-balance} characterizes a class of policies satisfying a property called diffusion regularity, which is defined in term of their induced maps.

% We introduce some other notations for 

% Notice single-arm relaxed problem (\ref{LP}) does not depend on $N$, so the complexity of solving the single-arm relaxed problem does not scale with $N$.

% This linear programming relaxation provides 
% Practically, Problem (\ref{KP}) is usually infeasible to be solved exactly

% Problem (\ref{KP}) may still be infeasible to solve in many cases since it is an infinite-horizon 

% It is usually not feasible to solve Problem (\ref{KP}) since its infinite
% This relaxation technique is useful in many aspects. First, since the set of feasible policies is augmented in the relaxation problem, its solution provides an upper bound for optimal policies' performance in the original problem (\ref{OP}). Second

% This relaxation provides many advantages.
% Noticing we drop the floor operator of $\lfloor \alpha N \rfloor$ in the relaxed version, 

\section{Diffusion Regularity Conditions}\label{sec-diff-balance}

This section defines a property called diffusion regularity and shows that policies possessing this property satisfy a bound on their corresponding diffusion statistics' first moments. This diffusion regularity property is shown to be satisfied by the fluid-balance policies proposed in section \ref{sect-6} and the bound shown here is a tool used to understand their performance theoretically in that section.

The intuition behind the diffusion regularity condition is that as long as $\tilde{X}_t^N$ remains bounded by a term that does not grow with $N$, then $\tilde{Z}_{t + 1}^N$ is also bounded by another term that does not grow with $N$. The first three conditions in the diffusion regularity condition are similar to conditions proposed in \cite{zhang2021restless} and the last condition is added specifically for our infinite-horizon setting to guarantee that diffusion statistics $\hat{Z}_t^N$ accumulate noise at no more than a linear rate over time.

% The insight behind diffusion regularity is, roughly speaking,
% as long as the diffusion statistic $\tilde{X}_t^N$ is bounded by $O(1)$, $\tilde{Z}_{t + 1}^N$ will also be bounded by $O(1)$.
% Thus, we could control the grow of the first moments of diffusion statistics $(\tilde{Z}_t^N, \tilde{X}_t^N)$ across periods (Lemma \ref{first-bound}).

We now define diffusion regularity.
\begin{definition}\label{def-diff}
A policy $\pi$ is {\it diffusion regular up to period $T$} if its induced maps $\tilde{\pi}_{t, N}$ satisfy the following conditions, where $|\cdot|$ is the Euclidean $L^1$-norm:
\begin{itemize}
  \item For any $t \le T$, there is a constant $C_1 > 0$ such that for all $N, \theta_1$ and $\theta_2$,
\begin{align*}
    |\tilde{\pi}_{t, N}(\theta_1) - \tilde{\pi}_{t, N}(\theta_2)| \leq C_1 |\theta_1 - \theta_2|;
\end{align*}
  \item For any $t \le T$, there is a constant $C_2 > 0$ such that for all $N$,
\begin{align*}
    |\tilde{\pi}_{t, N}(0)| \leq C_2;
\end{align*}
    \item For any $t \le T$, there is a map $\tilde{\pi}_{t, \infty}$ such that for all $\theta, \ \tilde{\pi}_{t, N}(\theta) \rightarrow \tilde{\pi}_{t, \infty}(\theta)$ as $N \rightarrow + \infty$;
    \item For any $t \le T$, there is a constant $C_3 > 0$ such that for all $N, \theta$ and $s \in S$, we have $$\sum_{a \in A} |\tilde{\pi}_{t, N}(\theta)(s, a)| \le |\theta(s)| + C_3.$$
\end{itemize}
\end{definition}

If a policy $\pi$ is diffusion regular up to period $T$, its diffusion statistics' first moments are bounded above by a linear function of time (Lemma \ref{first-bound}).
The proof of Lemma \ref{first-bound} may be found in the Appendix.

\begin{lemma} \label{first-bound}
If a policy $\pi$ is diffusion regular, then there exists constant $c_1$ and $c_2$ (neither depends on $T$), s.t. for all $t \le T$ and $N$ ($N$ could be infinity),
\begin{align*}
    \mathbb{E}[|\tilde{Z}_t^N|] \leq c_1 + c_2 t.
\end{align*}
\end{lemma}

% \begin{lemma} \label{moment-bound}
% If a policy $\pi$ is diffusion regular, then there exists an increasing sequence $\{c_t\}_{t = 1}^T$ (not depending on $T$), s.t. for all $t \le T$ and $N$ ($N$ could be infinity),
% \begin{align*}
%     \mathbb{E}[|\tilde{Z}_t^N|^2] \leq c_t.
% \end{align*}
% \end{lemma}

\section{Fluid-balance policy}\label{sect-6}
This section defines fluid-balance policies, and shows that all fluid-balance policies are diffusion regular and achieve an $O(\sqrt{N}) + O(N \gamma^T)$ opt gap.

Roughly speaking, a fluid-balance policy is parameterized by two components: an optimal solution of the LP relaxation and a prioritization scores over states.
The resulting fluid-balance policy pulls arms respecting two rules: a consistency rule and a prioritization rule. 
The consistency rule requires that the diffusion statistics $\tilde{X}_t^N(s, \cdot)$ share the same sign as $\tilde{Z}_t^N(s)$ for each state $s$. 
The prioritization rule requires pulling arms according to the prioritization score as much as possible while respecting the consistency rule.

Formally, a fluid-balance policy is parameterized by an occupation measure $\{x_t(s, a)\}_{t, s, a}$ solving truncated Problem ($\ref{LP}$) up to period $T$ and ``priority-score'' functions $\{ \mathcal{P}_t(\cdot) \}_{t \le T}$ assigning each state a time-dependent real number. 
The fluid-balance policy with these components is defined by Algorithm \ref{al:1}.

\begin{algorithm}
	\caption{Fluid-balance policy}\label{al:1}
	\hspace*{\algorithmicindent} \textbf{Input:} optimal occupation measure $(x_t(s, a))_{t \le T, s \in S, a \in A}$, priority-score functions $\{ \mathcal{P}_t \}_{t \le T}$.
	\begin{algorithmic}[1]
	\For {$t = 1, 2, ..., T$}\label{l:7}
		\State Input: the number $Z_t(s)$ of arms in state $s$ and their associated diffusion statistics $\tilde{Z}^N_t(s)$
		\State For each state $s$, set 
            $X_t(s, 1) \leftarrow \min\{Z_t(s), \lceil x_t(s, 1) N + \sqrt{N} |\tilde{Z}^N_t(s)| \rceil \}$
		\While {$\sum_s X_t(s, 1) > \lfloor \alpha_t N \rfloor$}
		\State Find the state $s$ with the lowest priority-score such that $$X_t(s, 1) > \max\{ 0,  \lfloor x_t(s, 1) N - \sqrt{N} |\tilde{Z}^N_t(s)| \rfloor\}$$
		\State $X_t(s, 1) \leftarrow X_t(s, 1) - 1$
		\EndWhile
	    \State For each state $s$, pull $X_t(s, 1)$ arms in state $s$
    \EndFor
	\end{algorithmic}
\end{algorithm}

We can show that any fluid-balance policy is diffusion regular and thus satisfies the bound on the expected $L^1-$ norm of its diffusion statistic provided in Lemma~\ref{first-bound}. Moreover, it actually satisfies a more explicit bound than Lemma \ref{first-bound}. These statements are shown in the following lemma, whose proof appears in the Appendix.
\begin{lemma}\label{th:3}
    Any fluid-balance policy $\pi$ is diffusion regular, and $\mathbb{E}_{\pi}[|\tilde{Z}_t^N|] \le 2 t |S|^2$ for $t \le T$.
\end{lemma}

Now we are able to show our main results
\begin{theorem}\label{prop-trunct-2}
Given any fluid-balance policy $\pi$, $V^*_N - V_N(\pi) = O(\sqrt{N}) + O(N \gamma^T)$.
\end{theorem}

\proof{Proof of Theorem \ref{prop-trunct-2}}
Denote $\pi^*$ the optimal policy maximizing the infinite-horizon Problem (\ref{OP}). Then by denoting $B := \max_{s, a} |r(s,a)|$,
\begin{align*}
V^*_N - V_N(\pi) 
&= \mathbb{E}_{\pi^*} \sum_{t = 1}^T \gamma^t r(s_{i, t}, a_{i, t} ) - \mathbb{E}_{\pi} \sum_{t = 1}^T \gamma^t r(s_{i, t}, a_{i, t} )
 + \mathbb{E}_{\pi^*} \sum_{t = T + 1}^{\infty} \gamma^t r(s_{i, t}, a_{i, t} ) - \mathbb{E}_{\pi} \sum_{t = T + 1}^{\infty} \gamma^t r(s_{i, t}, a_{i, t})\\
&\le \hat{V}_N(T) - \mathbb{E}_{\pi} \sum_{t = 1}^T \gamma^t r(s_{i, t}, a_{i, t}) + 2\sum_{t = T + 1}^{\infty} \gamma^t B N 
\end{align*}
where the last inequality is due to the definition of $\hat{V}_N(T)$. 

We deal with these two terms $\hat{V}_N(T) - \mathbb{E}_{\pi} \sum_{t = 1}^T \gamma^t r(s_{i, t}, a_{i, t})$ and $2\sum_{t = T + 1}^{\infty} \gamma^t B N $ separately. For the first term,
\begin{align*}
    \hat{V}_N(T) - \mathbb{E}_{\pi} \sum_{t = 1}^T \gamma^t r(s_{i, t}, a_{i, t})
    =-\sqrt{N} \mathbb{E}_{\pi} \sum_{t = 1}^T \sum_{s, a} \gamma^t r(s, a) \tilde{X}_{t}^N (s, a)
    \le \sqrt{N} B \mathbb{E}_{\pi} \left[ \sum_{t=1}^T\gamma^t |\tilde{Z}_t^N| \right].
\end{align*}
Recall Lemma \ref{th:3}, we have
\begin{align*}
    \hat{V}_N(T) - \mathbb{E}_{\pi} \sum_{t = 1}^T \gamma^t r(s_{i, t}, a_{i, t})
    \le \sqrt{N} B \mathbb{E}_{\pi} \left[ \sum_{t=1}^T\gamma^t |\tilde{Z}_t^N| \right]
    \le \sqrt{N} B \sum_{t=1}^T\gamma^t 2t|S|^2 \le \sqrt{N} \frac{2|S|^2\gamma B}{(1 - \gamma)^2} = O(\sqrt{N}).
\end{align*}

For the second term,
\begin{align*}
2\sum_{t = T + 1}^{\infty} \gamma^t B N
= \frac{2B \gamma }{1 - \gamma}N \gamma^T = O(N \gamma^T).
\end{align*}

Combining above analysis together, we conclude $V^*_N - V_N(\pi) = O(\sqrt{N}) + O(N \gamma^T)$.
\hfill $\Box$

Based on Theorem \ref{prop-trunct-2}, we can show the following proposition:
\begin{corollary}
Choosing $T = \frac{1}{2} \log_{\frac{1}{\gamma}} N$ implies that for any fluid-balance policy $\pi$,we have
\begin{equation}
    V_N^* - V_N(\pi) = O(\sqrt{N}).
\end{equation}
\end{corollary}

% \proof{Proof of Proposition \ref{prop-trunct-2}}
% \begin{align*}
% V^*_N - V_N(\pi) 
% &= \mathbb{E}_{\pi^*} \sum_{t = 1}^T \gamma^t r(s_{i, t}, a_{i, t} ) - \mathbb{E}_{\pi} \sum_{t = 1}^T \gamma^t r(s_{i, t}, a_{i, t} )\\
%  &+ \mathbb{E}_{\pi^*} \sum_{t = T + 1}^{\infty} \gamma^t r(s_{i, t}, a_{i, t} ) - \mathbb{E}_{\pi} \sum_{t = T + 1}^{\infthy} \gamma^t r(s_{i, t}, a_{i, t})\\
% &\le \hat{V}_N(T) - \mathbb{E}_{\pi} \sum_{t = 1}^T \gamma^t r(s_{i, t}, a_{i, t}) \\
% &+ 2\sum_{t = T + 1}^{\infty} \gamma^t B N\\
% &= -\sqrt{N} \mathbb{E}_{\pi} \sum_{t = 1}^T \sum_{s, a} \gamma^t r(s, a) \tilde{X}_{t}^N (s, a) + 2\sum_{t = T + 1}^{\infty} \gamma^t B N \\
% &\le \sqrt{N} B \mathbb{E}_{\pi} \left[ \sum_{t=1}^T\gamma^t |\tilde{Z}_t^N| \right] + 2B \gamma \frac{N \gamma^T}{1 - \gamma}\\
% &\le \sqrt{N} B \sum_{t=1}^T\gamma^t 2t|S|^2  + 2B \gamma \frac{N \gamma^T}{1 - \gamma}\\
% &\le 2|S|^2\sqrt{N} B \sum_{t = 1}^{\infty} t \gamma^t  + 2B \gamma \frac{N \gamma^T}{1 - \gamma}\\
% &\le \sqrt{N} \frac{2|S|^2\gamma B}{(1 - \gamma)^2}  + \frac{ 2B \gamma}{1 - \gamma} N \gamma^T
% \end{align*}

% Taking $T = \log_{\frac{1}{\gamma}} N$, we would have $V^*_N - V_N(\pi) \le \sqrt{N} \frac{2|S|^2\gamma B}{(1 - \gamma)^2}  + \frac{ 2B \gamma}{1 - \gamma}$.

\section{Numerical Experiment}\label{numerical}
This section illustrates the performance of fluid-balance policies through two numerical experiments, focusing on their advantage over the Whittle index policy. 

The first experiment studies a simple non-indexable example encapsulating a tradeoff that is important in many more complex real-world decision problems. The decision-maker can either generate a substantial reward over a short time horizon or generate a steady small amount of reward over a very long horizon. 
% We refer to this problem briefly as the Tortoise and the Hare, referencing Aesop's fable \pfcomment{Xiangyu to cite} tackling this tradeoff.
We refer to this as the {\it Slow-and-steady Problem}, borrowing from the idiom ``slow and steady wins the race''.
We show this example is not indexable, and thus the Whittle index is not well-defined. However, the fluid-balance policy can be computed analytically without truncation (i.e., the truncation point is $T=\infty$) and is actually the optimal policy.

In the second experiment, we compare the fluid-balance policy against the Whittle index for a discounted version of a problem studied in \cite{fu2019towards} and \cite{biswas2021learn}. Although this problem is indexable, the fluid-balance policy outperforms the Whittle index by over 30$\%$. 
This problem is drawn from the literature studying bandits with unknown transition kernels, where the Whittle index policy's performance is used to represent the performance achievable when transition kernels are known.
Our results suggest that the fluid-balance policy better represents the performance achievable with full information and thus might be a better benchmark than the Whittle index policy.

\subsection{The Slow-and-Steady Problem: Large Fleeting Rewards vs. Small Steady Rewards}
% \subsection{Fast-but-fleeting vs. Slow-but-steady}
% Consider the following example. 
This section constructs a simple non-indexable problem reflecting an important tradeoff arising in real-world decision making: should we generate one large reward immediately, or generate a sequence of small rewards over a much longer period. 
We refer to this as the ``slow-and-steady problem,'' echoing the proverb ``slow and steady wins the race'' often uttered when considering such tradeoffs. 
% We refer to these two policies as fast-but-fleeting and slow-but-steady.
We show that the Whittle index is not well defined for the slow-and-steady problem while the fluid-balance policy we consider is not just optimal asymptotically but also optimal for finite $N$.

\paragraph{Problem Definition} 
% \subsubsection{Problem Definition} 
% \label{sec:slow-steady-problem}
% C = End
% B* = End
% A' = Steady
% B' = Brief
% A = Pre-Steady
% B = Uncommitted-Brief
% A0 = Uncommitted-Steady
In the slow-and-steady problem there are two states that generate non-zero rewards: the ``Steady'' state and the ``Brief`` state. Once an arm enters the Steady state, it stays there and generates a reward of $\alpha \ge 0$ each time it is activated. An arm in the Brief state stays in the Brief state until it is activated, at which point it generates a reward of $\beta \ge 0$ and transitions to the ``End'' state. We think of $\alpha$ as being small and $\beta$ as being large. Once an arm is in the End state, it stays in that state.

Before transitioning into either the Steady or Brief states, an oscillates between the ``Uncommitted-Steady'' and ``Uncommitted-Brief'' states until the arm is activated. When activated, the arm transitions into the End state with a small probability $\epsilon$ and transitions into the Steady state (if it was in the Uncommitted-Steady state when it was activated) or the Brief state 
(if it was in the Uncommitted-Brief state when it was activated) otherwise.
For technical reasons, we also have a ``Pre-Steady'' state, which always transitions into the Steady state regardless of whether it was activated.

\begin{figure}
    \centering
    \includegraphics[scale=.7]{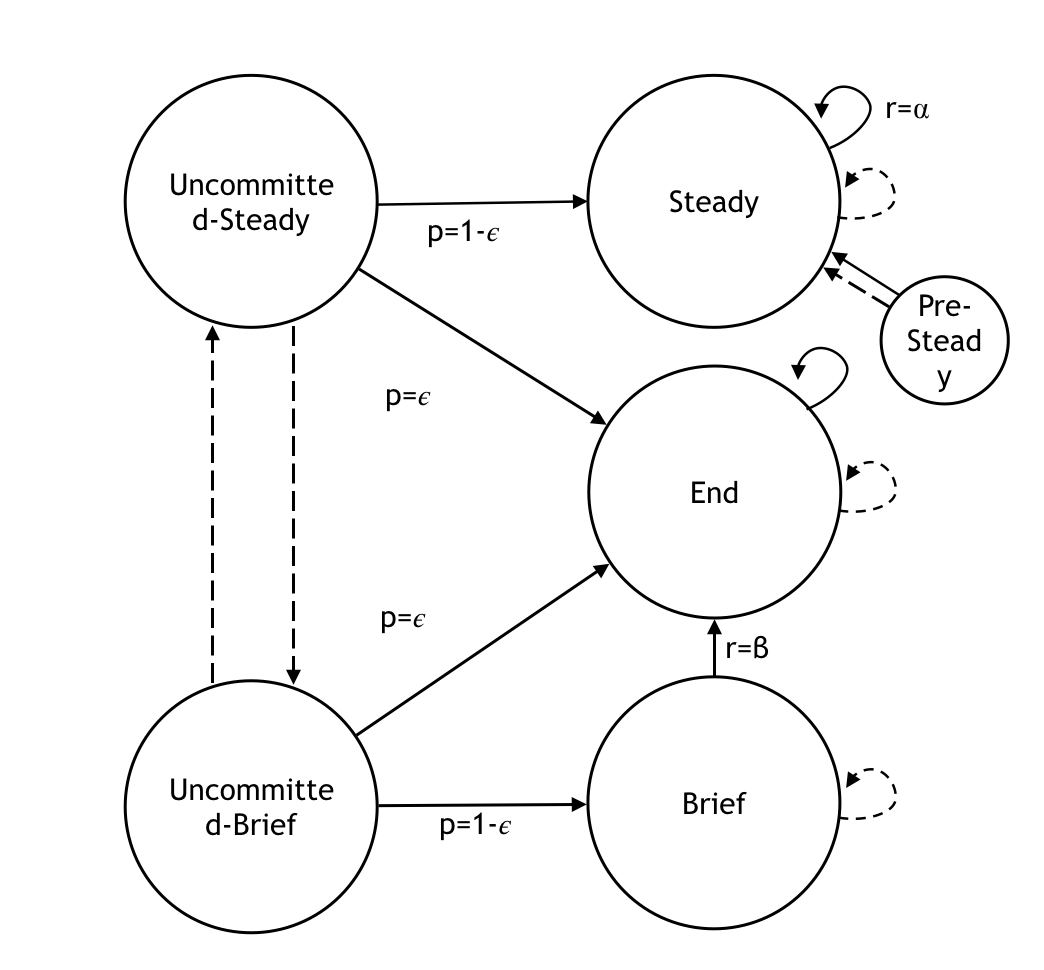}
    \caption{State transition diagram for Slow-and-Steady Problem. Each node stands for a state. Edges between nodes stand for transitions: dashed lines represent transitions when an arm is idled; solid lines represent transition when an arm is activated. Unless the transition probability is specified on the edge, it is 1. Similarly, unless the reward is specified on an edge, it is 0.}
    \label{slow-steady-diagram}
\end{figure}

Figure \ref{slow-steady-diagram} shows the transition dynamics between states, where nodes represent states and edges between nodes represent transition probabilities. Formally, the transition kernel is given by:
% In this problem, which we refer to as the ``Slow-and-steady Problem'', there are 7 different states: Pre-Steady, Uncommitted-Brief,Steady, Brief, T1 (Terimation 1), End 2(T2) Brief $\{ A_0, A, A', B, B', B^*, C\}$. The transition between different states are given as
\begin{align*}
    & \mathbb{P}[s_{t + 1} = \text{Steady}\ \mid\ s_{t} = \text{Steady} , a_t = a] = 1, \forall a \in \{ 0, 1 \},\\
    & \mathbb{P}[s_{t + 1} = \text{End}\ \mid\ s_{t} = \text{Brief} , a_t = 1] = 1, \quad
      \mathbb{P}[s_{t + 1} = \text{Brief}\ \mid\ s_{t} = \text{Brief} , a_t = 0] = 1,\\
    & \mathbb{P}[s_{t + 1} = \text{End}\ \mid\ s_{t} = \text{End} , a_t = a] = 1, \forall a \in \{ 0, 1 \},\\
    & \mathbb{P}[s_{t + 1} = \text{Uncommitted-Brief}\ \mid\ s_{t} = \text{Uncommitted-Steady}, a_t = 0] = 1,\\
    & \mathbb{P}[s_{t + 1} = \text{Uncommitted-Steady}\ \mid\ s_{t} = \text{Uncommitted-Brief}, a_t = 0] = 1.\\
    & \mathbb{P}[s_{t + 1} = \text{Steady}\ \mid\ s_{t} = \text{Uncommitted-Steady}, a_t = 1] = 1 - \epsilon,\\
    & \mathbb{P}[s_{t + 1} = \text{End}\ \mid\ s_{t} = \text{Uncommitted-Steady}, a_t = 1] = \epsilon, \\
    & \mathbb{P}[s_{t + 1} = \text{Brief}\ \mid\ s_{t} = \text{Uncommitted-Brief}, a_t = 1] = 1 - \epsilon, \\
    & \mathbb{P}[s_{t + 1} = \text{End}\ \mid\ s_{t} = \text{Uncommitted-Brief}, a_t = 1] = \epsilon,\\ 
    & \mathbb{P}[s_{t + 1} = \text{Steady}\ \mid\ s_{t} = \text{Pre-Steady}, a_t = a] = 1.
\end{align*}

% \begin{figure}
%     \centering 
%     \includegraphics[width=20cm, height=13cm]{InfiniteHorizonDiagram.pdf}
%     \label{slow-steady-diagram}
%     \caption{State transition diagram for Slow-and-Steady Problem}
% \end{figure}

Formally, the reward function is given by
\begin{align*}
    r(\text{Steady}, 1) = \alpha, \ r(\text{Brief}, 1) = \beta,
\end{align*}
with rewards for all other state-action pairs set to 0.
We seek to maximize the expected infinite-horizon discounted reward with discount factor $\gamma = 1 - \epsilon$. 

We assume our parameters satisfy $0\  < (1 + \frac{1}{\gamma^2}) \ \alpha \ < \ \beta \  < \ \frac{\gamma}{1 - \gamma} \alpha$, i.e. the Brief state generates a somewhat larger reward than Steady state, but not too much larger. Also we assume $\epsilon < \frac{1}{8}$, i.e. the Uncommitted-Brief and Uncommitted-Steady states transition into the Brief and Steady states with a reasonably high probability. In these parameter ranges, and for the budget and initial  occupation measure chosen below, we show below that the problem is not indexable.
% \pfcomment{can you add a sentence explaining why we make these assumptions?}

The budget is set so that we can pull  $\lfloor \gamma N \rfloor$ arms out of the $N$ total arms in each period. In the initial time period, $\lfloor \phi_1 N \rfloor$ arms in the \text{Uncommitted-Steady} state, $\lfloor \phi_2 N \rfloor$ arms are in the End state and $\lfloor \phi_3 N \rfloor$ arms are in the $\text{Pre-Steady}$ state, where $\phi_1 = 2 - \frac{1}{\gamma}, \ \phi_2 = \gamma + \frac{1}{\gamma} - 2$ and $\phi_3 = 1 - \gamma$. These parameters $\{ \phi_i \}_{i = 1, 2, 3}$ are chosen so that activating all arms in the Pre-Steady and Uncommitted-Steady states respects the constraint.

% At each time period, we can pull $\gamma N$ number of arms.

\paragraph{Analysis}

We first show, in the following proposition, that this problem is not indexable. Thus, the Whittle index is not well-defined. Proofs of this and other propositions in this section appear in the appendix.
\begin{proposition}\label{prop-1}
The slow-and-steady problem is not indexable.
\end{proposition}

Towards defining a fluid-balance policy, 
we show in the next proposition that the infinite-horizon linear programming relaxation of the slow-and-steady problem defined above permits an analytical solution. This allows defining fluid balance policies without truncation (i.e., the truncation point used is $T=\infty$).
\begin{proposition}\label{prop-2}
Consider the policy that pulls all arms in the $\text{Uncommitted-Steady}$ and $\text{Pre-Steady}$ in the first period, then pulls all arms in $Steady$ from the second period onward. This policy is optimal in the linear programming relaxation (\ref{KP_1}) of the slow-and-steady problem.
\end{proposition}

Using this optimal policy for the relaxed problem, we construct a fluid-balance policy. This fluid balance policy $\pi$ activates all arms in $\text{Uncommitted-Steady}$ and Pre-Steady states in the first period, then activates as many arms in the $\text{Steady}$ state as possible starting from the second period. If there are fewer than $\lfloor \gamma N \rfloor$ arms in the $\text{Steady}$ state, it activates arms in the $\text{End}$ state to meet the budget.

Theorem~\ref{prop-trunct-2} implies that the fluid-balance policy $\pi$ is asymptotically optimal, i.e., its opt gap is $O(\sqrt{N})$.
Surprisingly, this fluid-balance policy is not only asymptotically optimal, but optimal, i.e., its opt gap is $0$. This is shown in the following proposition.
\begin{proposition}\label{prop-3}
The fluid-balance policy $\pi$ is an optimal policy for the slow-and-steady problem.
\end{proposition}

The bound on the opt gap of a fluid-balance policy in Theorem \ref{prop-trunct-2} is derived by comparing a fluid-balance policy's reward expected total discounted reward against the optimal reward of the linear programming relaxation (which is an upper bound on the value of an optimal policy) rather than the value of an optimal policy in the original (unrelaxed) problem. Since the fluid-balance policy $\pi$ is optimal in the slow-and-steady problem, its opt gap is $0$. The optimal reward of the linear programming relaxation, however, is strictly bigger than that of an optimal policy, causing the gap to the linear programming relaxation to remain $O(\sqrt{N})$. This is shown in the following proposition.
\begin{proposition}\label{prop-4}
$\hat{V}_N^* - V_N(\pi) = \theta \sqrt{N} + O(1)$, where $\theta = \sqrt{\frac{(1 - \gamma) (2 \gamma - 1)}{2 \pi}}$.
\end{proposition}

\subsection{Bandit literature benchmark}
There is a stream of literature studying problems similar to the one we study, but where state transitions are unknown, e.g. \cite{fu2019towards} and \cite{biswas2021learn}.
Rather than designing policies based on knowledge of the problem's state transition kernel, as in the restless bandit problem we study, this stream of literature designs algorithms that estimate the state transition kernel while simultaneously choosing actions and collecting rewards.
A common practice in this literature is to benchmark a proposed algorithm's performance against a full-information policy, such as the Whittle index policy, on a specific problem.

In this section, we study a problem based on one of the most commonly used benchmark problems from this literature. This problem is based on \cite{fu2019towards} and \cite{biswas2021learn}, who study an undiscounted version in which $N/10$ arms are pulled per period.
In the setting where the budget is $N/10$ arms are pulled per period, a numerical study shows that the Whittle index is actually also a fluid balance policy, and thus is asymptotically optimal by Theorem~\ref{prop-trunct-2}. When the budget is $N/2$ arms per period, however, we find that the fluid-balance policy outperforms the Whittle index by over $30\%$, suggesting that fluid-balance policies provide better full-information benchmark.

\paragraph{Problem Definition} 

In the problem that we study, there are 4 different states: $\{0, 1, 2, 3\}$. Transition kernels $P_a = {p(s, 0, s')}_{s, s' \in S}$ for action $a = 0$ and $a = 1$ are given by
$$P_0 = 
\begin{pmatrix}
    1/2 &  \ 0  & \ 0  &\ 1/2\\
    1/2 &\ 1/2 &\ 0 &\ 0\\
    0 &\ 1/2 &\ 1/2 &\ 0\\
    0 &\ 0 &\ 1/2 &\ 1/2
\end{pmatrix}, 
\ 
P_1 = 
\begin{pmatrix}
    1/2 &  \ 1/2  & \ 0  &\ 0\\
    0 &\ 1/2 &\ 1/2 &\ 0\\
    0 &\ 0 &\ 1/2 &\ 1/2\\
    1/2 &\ 0 &\ 0 &\ 1/2
\end{pmatrix}.
$$

The reward solely depends on the state and is unaffected by the action: $$r(0, a) = -1, \ r(1, a) = 0, \ r(2, a) = 0, \ r(3, a) = 1; \forall a \in \{0, 1\}.$$

We set the discount factor to $\gamma = 1/2$ and require $N/2$ arms to be pulled per period.
Initially, there are $N / 6$ arms in state $0$, $N / 3$ arms in state $1$ and $N / 2$ arms in state $2$.

\paragraph{Analysis} 
% \textbf{Indexability} 
Via direct calculation, we can show that this problem is indexable. Ranking states from the highest to the lowest according to the Whittle index, we have that the Whittle index prioritizes state 2 over state 1, and state 1 over state 3.
It is unknown whether the Whittle index policy is asymptotically optimal in this problem, though numerical experiments below suggest that it is not.

% \textbf{Fluid-balance policy} 
To compute the fluid-balance policy,
Since this infinite-horizon problem's linear programming relaxation does not permit an analytical solution, we solve the truncated version up to $T = 100$. This provides an accurate approximation of the upper bound implied by the linear programming relaxation because the total reward after period $100$ is less than $2^{-100}$, much smaller than the precision $10^{-17}$ of a 64-bit floating point number.

After solving the truncated relaxation problem, we need a final piece before implementing a fluid-balance policy: the prioritization score. We adhere to Whittle index here, where we rank states from high to low as state 2 $>$ state 1 $>$ state 0 $>$ state 3.

\begin{figure}[h]
	\centering
	\subfloat[]{\includegraphics[width=8cm, height = 6 cm]{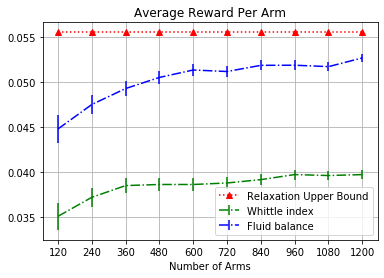}\label{fig:f1}}
	\hfill
	\subfloat[]{\includegraphics[width=8cm, height = 6 cm]{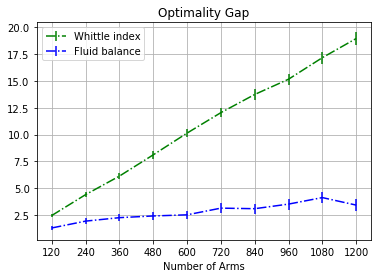}\label{fig:f2}}
	\caption{\label{fig:experiment}
	Performance comparison between Whittle index and fluid-balance policy.
	The left panel shows the average reward per arm versus number of arms ($N$), where we compare the upper bound from the linear programming relaxation with the performance of the Whittle index and fluid-balance policies as estimated via simulation.
	The right panel shows an upper bound on the opt gap (the upper bound from the linear programming relaxation minus a simulation-based estimate of the expected total discounted reward) versus the number of arms ($N$). The Whittle index's opt gap grows linearly with $N$, consistent with a lack of asymptotic optimality, while the fluid-balance policy's opt gap grows sublinearly, consistent with asymptotic optimality and our result from Theorem~\ref{prop-trunct-2} that its opt gap is $O(\sqrt{N})$.}
\label{crowdsource}
\end{figure}

Figure~\ref{fig:experiment} compares the performance of the Whittle index and fluid-balance policies.
Simulation with 2000 independent replications is used to estimate the performance of each policy via the sample mean and $95\%$ confidence intervals. As we can see, the fluid-balance policy outperforms the Whittle index in both the small-$N$ and large-$N$ regimes.

\section{Conclusion and Future Work}\label{conclusion}
In this paper, we propose a class of policies, called fluid-balance policies, which achieve an $O(\sqrt{N})$ opt gap universally as the number of arms $N$ grows large. Unlike the Whittle index policy, fluid-balance policies do not require an indexability condition to be well-defined and our results show they are asymptotically optimal without the need for difficult-to-verify sufficient conditions.

Although we restrict our analysis to restless bandits, we believe the techniques and insights we develop here can be generalized to multi-action multi-resource restless bandits, also known as weakly coupled Markov Decision Processes. Another interesting direction for future work would be characterizing a non-degeneracy condition, similar to \cite{zhang2021restless}, sufficient for a fluid balance policy to achieve a $O(1)$ optimality gap in an infinite-horizon restless bandit.

\bibliographystyle{informs2014}
\bibliography{References}
%Through out all the proof here, we assume $N$ is taken s.t $\alpha_t N$ and $m_s N$ are always integer valued.
{\section{Appendix}}

This section provides all technical proof in the main paper.
\subsection{Proof for Lemma \ref{fenchel}}\label{proof-fenchel}
In original problem (\ref{OP}) the budget constraint is in sense of cardinality, while the expectation constraint is need for relaxation problem (\ref{KP}). So a wider class of policy is feasible in the relaxation problem, which implies
\begin{align}
    V_N^*(T) \leq \hat{V}_N^{*}(T).
\end{align}
To prove $\hat{V}^*_N(T) = N \hat{V}_1^*(T)$, we use Lagrangian Relaxation similar to \cite{farias2011irrevocable, guha2008sequential} as the key idea in the following argument.

Through imitating straightforwardly the proof of Fenchel Duality Theorem  \citep{rockafellar1970convex},
\begin{align}\label{eq-8}
    \max_{\pi} \min_{\lambda} \mathbb{E}_{\pi} \sum_{t = 1}^T \gamma^t R(\mathbf{s}_t,\mathbf{a}_t) + \lambda_t (\alpha_t N - |\mathbf{a}_t|) = \min_{\lambda} \max_{\pi} \mathbb{E}_{\pi} \sum_{t = 1}^T \gamma^t R(\mathbf{s}_t,\mathbf{a}_t) + \lambda_t (\alpha_t N - |\mathbf{a}_t|)
\end{align}
where $\lambda = (\lambda_1, \lambda_2, ..., \lambda_T)$.

The let-hand side of Equation (\ref{eq-8}) equals to $\hat{V}_N^*(T)$. On the right hand side, for fixed $\lambda$,
\begin{align*}
    \mathbb{E}_{\pi} \sum_{t = 1}^{T} \gamma^t R(\mathbf{s}_t,\mathbf{a}_t) + \lambda_t (\alpha_t N - |\mathbf{a}_t|)
    = \mathbb{E}_{\pi} \sum_{t = 1}^{T} \sum_{i = 1}^N \gamma^t r(s_{t, i}, a_{t, i}) + \lambda_t(\alpha_t - a_{t, i}).
\end{align*}
Since all arms share the same transition kernel and reward function, 
\begin{align*}
    \mathbb{E}_{\pi} \sum_{t = 1}^{T} \sum_{i = 1}^N \gamma^t r(s_{t, i}, a_{t, i}) + \lambda_t(\alpha_t - a_{t, i}) = N \ \mathbb{E}_{\pi} \sum_{t = 1}^{T} \gamma^t r(s_{t, 1}, a_{t, 1}) + \lambda_t( \alpha_t - a_{t, 1}).
\end{align*}
So we conclude
\begin{align}\label{eq-9}
  \min_{\lambda} \max_{\pi} \mathbb{E}_{\pi} \sum_{t = 1}^T \gamma^t R(\mathbf{s}_t,\mathbf{a}_t) + \lambda_t (\alpha_t N - |\mathbf{a}_t|)
    =  N \min_{\lambda} \max_{\pi} \mathbb{E}_{\pi} \sum_{t = 1}^{T} \gamma^t r(s_{t, 1}, a_{t, 1}) + \lambda_t(\alpha_t - a_{t, 1}).
\end{align}

By using Fenchel Duality again on the one-arm problem,
\begin{align}\label{eq-10}
    \hat{V}_1^*(T) 
    &= \max_{\pi} \min_{\lambda} \mathbb{E}_{\pi} \sum_{t = 1}^T \gamma^t r(s_{t, 1}, a_{t, 1}) + \lambda_t(\alpha_t - a_{t, 1}) \nonumber \\
    &= \min_{\lambda} \max_{\pi} \mathbb{E}_{\pi} \sum_{t = 1}^T \gamma^t r(s_{t, 1}, a_{t, 1}) + \lambda_t(\alpha_t - a_{t, 1}).
\end{align}

To summarize Equation (\ref{eq-8}), (\ref{eq-9}) and (\ref{eq-10}) together,
\begin{align*}
    \hat{V}_N^*(T) = N \ \hat{V}_1^*(T).
\end{align*}

% \section{Proof of Theorems in Section \ref{gap}}
% We prove all the theoretical results in Section \ref{gap} here.

\subsection{Discussion of the rounding error in budget constraints}\label{al:2-1}
We want to show a rounding error in the relaxation Problem (\ref{KP}) results in at most a constant difference in the optimal objective value. Mathematically speaking, denote
\begin{align*}
    \hat{V}_N^*(T) &= \max_{\pi} \left\{ \mathbb{E}_{\pi} \sum_{t=1}^{T} \gamma^t R(\mathbf{s}_t, \mathbf{a}_t) \Bigg| \mathbb{E}|\mathbf{a}_t| = \alpha_t N, \ \forall t \le T \right\},  \\
    \hat{V}_{N, R}^*(T) &= \max_{\pi} \left\{ \mathbb{E}_{\pi} \sum_{t=1}^{T} \gamma^t R(\mathbf{s}_t, \mathbf{a}_t) \Bigg| \mathbb{E}|\mathbf{a}_t| = \lfloor \alpha_t  N\rfloor, \ \forall t \le T \right\}.
\end{align*}
Then $|\hat{V}_{N}^* - \hat{V}_{N, R}^*| \le c$, where $c$ does not depend on $N$. Thus, all our analysis on the asymptotic regime of opt gap holds true since the LP relaxation upper bound (in rounded version) deviates from the unrounded version at most a constant away, not affecting the asymptotic analysis.

The proof of the above statement is straight forward. As seen from Lemma \ref{fenchel}, there exists a single-arm pulling strategy which pulls $\alpha_t$ arms per period in expectation and achieves objective value $\hat{V}_{1}^*$. Thus, we can pull $N - 1$ arms according to this strategy and pull the only arm left with probability $\lfloor \alpha_t N \rfloor - \alpha_t (N - 1)$ at period $t$. Thus, we show
\begin{align*}
    \frac{N - 1}{N}  \hat{V}_{N}^*(T) - \hat{V}_{N, R}^*(T) \le \ \frac{\max_{s, a} r(s, a)}{1 - \gamma}.
\end{align*}
Similarly, we can show
\begin{align*}
    \frac{N - 1}{N}  \hat{V}_{N, R}^*(T) - \hat{V}_{N}^*(T) \le \ \frac{\max_{s, a} r(s, a)}{1 - \gamma}.
\end{align*}

Combining the above two inequality concludes the statement.

\subsection{Proof of Lemma \ref{first-bound}}

To prove Lemma \ref{first-bound}, first notice
\begin{equation}\label{dynamic}
    Z^N_{t+1}(s) = \sum_{s' \in S, a \in A} \sum_{i=1}^{N} 1(s_{t, i} = s', a_{t, i} = a, s_{t + 1, i} = s)
\end{equation}
where $1(s_{t, i} = s', a_{t, i} = a, s_{t + 1, i} = s)$ is the indicator function of event $\{s_{t, i} = s', a_{t, i} = a, s_{t + 1, i} = s\}$.
By dynamic equation (\ref{dynamic}) in a vector form,
\begin{align*}
    Z_{t+1}^N = \sum_{s' \in S, a \in A} B_t^N(s', a),
\end{align*}
where $B_t^N(s', a)$ is the sum of $X_t^N(s', a)$ independent $|S|$-dimensional Bernoulli random variable with mean $(p(s', a, s))_{s \in S}$. For simplicity, we denote $p(s', a) := (p(s', a, s))_{s \in S}$ in the following proof.
With this new notation, 
\begin{align*}
    B_t^N(s', a) \sim \text{Binomial}(X_t^N(s', a), p(s', a)).
\end{align*}

Recall that $X_t^N(s', a)$ can be decomposed as $N x_t(s', a) + \sqrt{N} \tilde{X}_t^N(s', a)$. According to Lemma \ref{decompose-bin}, there exists two random variables $C_t^N(s', a)$ and $\Delta_t^N(s', a)$, s.t.
\begin{align}\label{decomp-eq}
    B_t^N(s', a) = C_t^N(s', a) + \Delta_t^N(s', a),
\end{align}
and that, conditionally on $X_t^N(s', a)$, have marginal distribution:
\begin{align*}
    &C_t^N(s', a) \sim \text{Binomial}( \lceil N x_t(s', a) \rceil, p(s', a)),\\
    &\Delta_t^N(s', a) \sim sgn(\tilde{X}_t^N(s', a)) \text{Binomial}(\lfloor \sqrt{N} \left| \tilde{X}_t^N(s', a) \right| \rfloor , p(s', a)).
\end{align*}

By equation (\ref{decomp-eq}) and recall the definition of diffusion statistics,
\begin{align*}
    \tilde{Z}_{t + 1}^N = 
    &\frac{1}{\sqrt{N}} \sum_{s' \in S, a \in A} C_t^N(s', a) - x_t(s', a) p(s', a) N + \frac{1}{\sqrt{N}} \sum_{s' \in S, a \in S} \Delta_t^N(s', a).
\end{align*}

Now consider the $L^{1}$-norm of $\tilde{Z}_t^N$,
\begin{align*}
    |\tilde{Z}_{t + 1}^N|
    &\leq \Big|\frac{1}{\sqrt{N}} \sum_{s' \in S, a \in S} \Delta_t^N(s', a)\Big| + \frac{1}{\sqrt{N}} \sum_{s' \in S, a \in A}  \Big| C_t^N(s', a) - x_t(s', a) p(s', a) N\Big|\\
    &\leq \sum_{s \in S, a \in A}| \tilde{X}_t^N(s', a)| + \frac{1}{\sqrt{N}} \sum_{s' \in S, a \in A} \Big| C_t^N(s', a) - x_t(s', a) p(s', a) N\Big|\\
    &\leq |\tilde{Z}_t^N| + \frac{C_3|S|}{\sqrt{N}} + \frac{1}{\sqrt{N}} \sum_{s' \in S, a \in A} \Big| C_t^N(s', a) - x_t(s', a) p(s', a) N\Big|,
\end{align*}
where the last inequality is due to diffusion regularity of the policy so that $\sum_{a \in A} |\tilde{X}_t^N(s, a)| \le |\tilde{Z}_t^N(s)| + \frac{C_3}{\sqrt{N}}$.
Thus,
\begin{align*}
    \mathbb{E} |\tilde{Z}_{t + 1}^N|
    &= \mathbb{E} |\tilde{Z}_t^N| + \frac{C_3|S|}{\sqrt{N}} + \frac{1}{\sqrt{N}} \sum_{s' \in S, a \in A} \mathbb{E} \Big| C_t^N(s', a) - x_t(s', a) p(s', a) N\Big| \\
    &\leq \mathbb{E} |\tilde{Z}_t^N| + \frac{C_3|S|}{\sqrt{N}} + \frac{1}{\sqrt{N}} \sum_{s' \in S, a \in A} \mathbb{E} \Big| C_t^N(s', a) - x_t(s', a) p(s', a) N\Big| \\
    &\leq \mathbb{E} |\tilde{Z}_t^N| + \frac{C_3|S|}{\sqrt{N}} + \frac{1}{\sqrt{N}} \sum_{s' \in S, a \in A, s \in S} \sqrt{\frac{1 + N x_t(s', a)}{4}},
\end{align*}
where the last inequality is by Lemma \ref{l1-bound}.

So 
\begin{align*}
    \mathbb{E} |\tilde{Z}_{t + 1}^N|
    &\leq \mathbb{E} |\tilde{Z}_t^N| + \frac{C_3|S|}{\sqrt{N}} + \frac{1}{\sqrt{N}} \sum_{s' \in S, a \in A, s \in S} \sqrt{\frac{1 + N x_t(s', a)}{4}} \\
    &= \mathbb{E} |\tilde{Z}_t^N| + C_3|S| + \frac{1}{2} \sum_{s' \in S, a \in A, s \in S} \sqrt{x_t(s', a) + 1} \\
    &\leq \mathbb{E} |\tilde{Z}_t^N| + C_3|S| + 2|S|^2.
\end{align*}

Thus, by induction we have 
\begin{align}\label{moment-linear-bound}
\mathbb{E} |\tilde{Z}_{t + 1}^N| \leq t (2|S|^2 +C_3|S|) +  \sup_N \mathbb{E} |\tilde{Z}_{1}^N|
\end{align}
Taking $c_1 = \sup_N \mathbb{E} |\tilde{Z}_{1}^N|$ and $c_2 = 2|S|^2 +C_3|S|$ concludes the proof.
\hfill $\Box$

We state and prove the following Lemma \ref{decompose-bin} and \ref{l1-bound}.
\begin{lemma}\label{decompose-bin}
Suppose random variable $S$ is a Binomial random variable with parameter $n$ and $p$, i.e., distributed as the sum of $n$ i.i.d. Bernoulli r.v.s with mean $p$.
Then for a given non-negative integer $m$, there exists random variable $S_1$ and $S_2$, s.t.
$S = S_1 + S_2$, and
\begin{align*}
    S_1 \sim \text{Binomial}(m, p), \ S_2 \sim sgn(n - m) \text{Binomial}(|n - m|, p),
\end{align*}
where $sgn(\cdot)$ is the sign function.
\end{lemma}

\proof{Proof of Lemma \ref{decompose-bin}}
There exists a sequence of i.i.d random variables $X_i \sim \text{Binomial}(1, p)$, s.t. 
$$S = \sum_{i = 1}^n X_i.$$
If $n > m$, taking $S_1 = \sum_{i=1}^m X_i, S_2 = \sum_{j = m + 1}^m X_j$ concludes the proof. If $n \leq m$, taking $S_1 = \sum_{i=1}^m X_i, S_2 = -\sum_{j = n + 1}^m X_j$ concludes the proof.
\hfill $\Box$

\begin{lemma}\label{l1-bound}
Suppose there are $n$ i.i.d Bernoulli random variable $X_1, X_2, ..., X_n$ with mean $p$. Then
\begin{align*}
    \mathbb{E} |\sum_{i = 1}^n X_i - n p| \leq \sqrt{\frac{n}{4}}.
\end{align*}
\end{lemma}

\proof{Proof of Lemma \ref{l1-bound}}
Direct calculation shows
\begin{align*}
    \mathbb{E} |\sum_{i = 1}^n X_i - n p| \leq \sqrt{\mathbb{E} |\sum_{i = 1}^n X_i - n p|^2} = \sqrt{np(1 - p)} \le \sqrt{\frac{n}{4}}.
\end{align*}

\subsection{Proof of Lemma \ref{th:3}}
Given a fluid-balance policy $\pi$, we directly check the induced map $\tilde{\pi}_{t, N}$ satisfies all three conditions in Definition \ref{def-diff}.

\proof{Verification of Condition 1}
Write the induced map in the component form $\tilde{\pi}_{t, N} = (\tilde{\pi}_{t, N}^1, ..., \tilde{\pi}_{t, N}^{|S|})$, and a direct calculation shows each component function $\tilde{\pi}_{t, N}^i \ (1 \leq i \leq |S|)$ is continuous, piece-wise linear, and has bounded gradient when exits. Mathematically speaking, there exits a constant $\tilde{C}_1$, s.t., for any $\theta$, any $t$ and any $N$,
\begin{align*}
    | \nabla \tilde{\pi}_{t, N}^i (\theta) | \leq \tilde{C}_1, \text{\ when $\nabla \tilde{\pi}_{t, N}^i (\theta)$ exits.}
\end{align*}

For any $\theta_1$ and $\theta_2$, there exits a sequence $(\theta_{1, 2}^0, \theta_{1, 2}^1, ..., \theta_{1, 2}^m)$ lies on the line segment between $\theta_1$ and $\theta_2$, s.t.

1. $\tilde{\pi}_{t, N}^i$ restricted on line segment between $\theta_{1, 2}^j$ and $\theta_{1, 2}^{j + 1}$ is linear for $j = 0, 1, ..., m - 1$

2. $\theta_{1, 2}^0 = \theta_1$ and $\theta_{1, 2}^m = \theta_2$.

Thus
\begin{align*}
    |\tilde{\pi}_{t, N}^i (\theta_1) - \tilde{\pi}_{t, N}^i (\theta_2)| \leq \sum_{j = 0}^{m - 1} |\tilde{\pi}_{t, N}^i (\theta_{1, 2}^j) - \tilde{\pi}_{t, N}^i (\theta_{1, 2}^{j + 1})|
    \leq \sum_{j = 0}^{m - 1} \tilde{C}_1 |\theta_{1, 2}^j - \theta_{1, 2}^{j + 1}|
    = \tilde{C}_1 |\theta_1 - \theta_2|.
\end{align*}

So by taking $C_1 = |S| \tilde{C}_1$,
\begin{align*}
    |\tilde{\pi}_{t, N} (\theta_1) - \tilde{\pi}_{t, N} (\theta_2)| \leq \sum_{i = 1}^{|S|} |\tilde{\pi}_{t, N}^i (\theta_1) - \tilde{\pi}_{t, N}^i (\theta_2)|
    \leq \sum_{i=1}^{|S|} \tilde{C}_1 |\theta_1 - \theta_2| = C_1 |\theta_1 - \theta_2|. 
\end{align*}

\proof{Verification of Condition 2}
Direct calculation shows $\tilde{\pi}_{t, N} (0) = 0$.

\proof{Verification of Condition 3}
Direct calculation shows $\tilde{\pi}_{t, \infty} (\tilde{Z}_{t}^{\infty})$ is a piece-wise linear map. 

\proof{Verification of Condition 4}
Direct calculation shows $\tilde{X}_{t, N}(s, 0)$ and $\tilde{X}_{t}^{N}(s, 1)$ has the same sign with $\tilde{Z}_{t}^{N}(s)$. 
% {Xiangyu provide more details}

To summarize,  we prove the induced map of any fluid-balance policy satisfies all four conditions in Definition \ref{def-diff}. Thus, any fluid-balance policy is diffusion regular.

As for the moment bound, notice $\tilde{Z}_{1}^N = 0$ and $C_3 = 0$ for any fluid balance policy. Thus, direct calculation of the coefficient in  (\ref{moment-linear-bound}) gives $\mathbb{E}_{\pi}[|\tilde{Z}_t^N|] \le 2 t |S|^2$ for $t \le T$.
\hfill $\Box$

\subsection{Proof of Proposition \ref{prop-1}}\label{app-prop-1}
We show that
\begin{itemize}
    \item with Lagrangian penalty $\lambda = 0$, state $A$ is active while state $B$ is inactive;
    \item with Lagrangian penalty $\lambda = \alpha$, state $A$ is inactive while state $B$ is active. 
\end{itemize}
Thus, the problem is not indexable. Now we analyze these two cases separately.

To be consistent with \cite{whittle1980multi}, we denote $V_\lambda(x)$ for the reward-to-go function of initial state $x$ with Lagrangian penalty $\lambda$.

With Lagrangian penalty $\lambda = \alpha$, we can see $V_\lambda(A') = 0, V_\lambda(C) = 0, V_\lambda(B^*) = 0$ and $V_\lambda(B') = \beta - \alpha$ via direct calculation. Thus, we have
\begin{align*}
    V_\lambda(A) &= \max \{ \gamma V_\lambda(B), -\alpha \},\\
    V_\lambda(B) &= \max \{ \gamma V_\lambda(A), -\alpha + \gamma (1 - \epsilon) V_\lambda(B') \}.
\end{align*}
Solving the above equations gives us
\begin{align*}
    V_\lambda(A) = \gamma (\gamma^2 (\beta - \alpha) - \alpha), V_\lambda(B) = \gamma^2 (\beta - \alpha) - \alpha.
\end{align*}
Thus, state $A$ is inactive and state $B$ is active.

With Lagrangian penalty $\lambda = 0$, we can see $V_\lambda(A') = \frac{\alpha}{1 - \gamma}, V_\lambda(C) = 0, V_\lambda(B^*) = 0$ and $V_\lambda(B') = \beta$ via direct calculation. Thus, we have
\begin{align*}
    V_\lambda(A) &= \max \{ \gamma V_\lambda(B), (1 - \epsilon) \gamma V_\lambda(A') \},\\
    V_\lambda(B) &= \max \{ \gamma V_\lambda(A), \gamma (1 - \epsilon) V_\lambda(B') \}.
\end{align*}
Solving the above equations gives us
\begin{align*}
    V_\lambda(A) = \frac{\alpha \gamma^2}{1 - \gamma}, V_\lambda(B) = \frac{\alpha \gamma^3}{1 - \gamma}.
\end{align*}
Thus, state $A$ is active and state $B$ is inactive.

\subsection{Proof of Proposition \ref{prop-2}}\label{app-prop-2}
Regardless of the budget constraint, 
\begin{itemize}
    \item the maximal reward generated from an initial state $A_0$ is up bounded by $\frac{\gamma \alpha}{1 - \gamma}$,
    \item the maximal reward could be generated from an initial state $A$ is up bounded by $(1 - \epsilon) \frac{\gamma \alpha}{1 - \gamma}$.
\end{itemize}
Thus, the maximal reward generated from $N$ arms is up bounded by
$$\phi_1 N (1 - \epsilon) \frac{\gamma \alpha}{1 - \gamma} + \phi_3 N \frac{\gamma \alpha}{1 - \gamma} = \frac{\gamma \alpha}{1 - \gamma} \gamma N.$$

Now we calculate the reward generated by policy $\pi$ which pulls all arms in $A_0$ and $A$ in the first period, and then pulls all arms in $A'$ starting from the second period.

In the first period, zero reward is generated.
In the second period, there are $\phi_3 N + \phi_1 N (1 - \epsilon) = \gamma N$ arms in state $A'$ in expectation. Starting from second period, we pull all arms in state $A'$. Thus, we generate $\gamma N * \frac{\alpha}{1 - \gamma} \gamma = \frac{\gamma \alpha}{1 - \gamma} \gamma N$ amounts of reward.

\subsection{Proof of Proposition \ref{prop-3}}\label{app-prop-3}
We only need to show that the optimal policy pulls all arms in state $A$ in the first period. 

If not, then under the optimal policy $\pi^*$, there exists an arm $x$ in state $A$ being idled and an arm $y$ in state $A_0$ or $C$ being pulled in the first period. Now we construct a policy $\hat{\pi}$ which pulls arm $x$ in the first period with satisfying $V_N(\hat{\pi}) \ge V_N(\pi^*)$.

The definition of $\hat{\pi}$ is pretty simple. It pull $x$ and idles $y$ in the first period, and starting from the second period, $\hat{\pi}$ takes the exact same action with $\pi$ for every arm. Then for any trajectory $\omega$, we can calculate the reward generated from arm $x$ by $\pi^*$ and $\hat{\pi}$ and show $V_N(\hat{\pi}) \ge V_N(\pi^*)$.

\subsection{Proof of Proposition \ref{prop-4}}\label{app-prop-4}
Under fluid-balance policy $\pi$, $\hat{V}_N^* - V_N(\pi) = \frac{\gamma \alpha}{1 - \gamma} \mathbb{E} [\gamma N - \min\{Z, \gamma N\}]$, where $Z$ is number of arms in state $A'$ starting from the second period. 

Thus, we have
$$Z = (1 - \gamma) N + \sum_{i = 1}^{(2 - \frac{1}{\gamma}) N} I_i$$
where $I_i$ is a Bernoulli r.v. with probability $\gamma$. Thus,
\begin{align*}
    \mathbb{E} [\gamma N - \min\{Z, \gamma N\}] = \mathbb{E} \left[ \sum_{i = 1}^{(2 - \frac{1}{\gamma}) N} I_i - \gamma \right]^+.
\end{align*}

By concentration inequality, we know $\mathbb{E} [\gamma N - \min\{Z, \gamma N\}] = \theta \sqrt{N} + O(1)$ where $\theta = \sqrt{\frac{(1 - \gamma) (2 \gamma - 1)}{2 \pi}}$.

\end{document}